\documentclass[lettersize,journal]{IEEEtran}
\usepackage{amsmath,amsfonts}
\usepackage{algorithmic}
\usepackage{algorithm}
\usepackage{array}
\usepackage[caption=false,font=normalsize,labelfont=sf,textfont=sf]{subfig}
\usepackage{textcomp}
\usepackage{stfloats}
\usepackage{url}
\usepackage{verbatim}
\usepackage{lipsum}
\usepackage{graphicx}
\usepackage{booktabs}
\usepackage{bbding}
\usepackage{color}
\usepackage{cite}
\usepackage{multirow}
\newcommand{\etc}{\textit{etc}. }
\newcommand{\ie}{\textit{i}.\textit{e}., }

\newcommand{\eg}{\textit{e.g., }}

\begin{document}

\title{Visual Commonsense-aware Representation Network for Video Captioning}

\author{Pengpeng Zeng, Haonan Zhang, Lianli Gao, Xiangpeng Li, Jin Qian and Heng Tao Shen,~\IEEEmembership{Fellow,~IEEE}
\thanks{Pengpeng Zeng, Haonan Zhang, Lianli Gao, Xiangpeng Li and Heng Tao Shen are with the
Future Media Center and School of Computer Science and Engineering, University of Electronic Science and Technology of China, Chengdu,
China, 611731. Jin Qian is with Southwest Jiaotong University.
\\
Corresponding author: Lianli Gao. E-mail: lianli.gao@uestc.edu.cn}}


\markboth{Journal of \LaTeX\ Class Files,~Vol.~14, No.~8, August~2021}%
{Shell \MakeLowercase{\textit{et al.}}: A Sample Article Using IEEEtran.cls for IEEE Journals}

\IEEEpubid{0000--0000/00\$00.00~\copyright~2021 IEEE}

\maketitle

\begin{abstract}
Generating consecutive descriptions for videos, \ie Video Captioning, requires taking full advantage of visual representation along with the generation process. Existing video captioning methods focus on making an exploration of spatial-temporal representations and their relationships to produce inferences. However, such methods only exploit the superficial association contained in the video itself without considering the intrinsic visual commonsense knowledge that existed in a video dataset, which may hinder their capabilities of knowledge cognitive to reason accurate descriptions. To address this problem, we propose a simple yet effective method, called Visual Commonsense-aware Representation Network (VCRN), for video captioning. Specifically, we construct a Video Dictionary, a plug-and-play component, obtained by clustering all video features from the total dataset into multiple clustered centers without additional annotation. Each center implicitly represents a visual commonsense concept in the video domain, which is utilized in our proposed Visual Concept Selection (VCS) to obtain a video-related concept feature. Next, a Conceptual Integration Generation (CIG) is proposed to enhance the caption generation. Extensive experiments on three publicly video captioning benchmarks: MSVD, MSR-VTT, and VATEX, demonstrate that our method reaches state-of-the-art performance, indicating the effectiveness of our method. In addition, our approach is integrated into the existing method of video question answering and improves this performance, further showing the generalization of our method. Source code has been released at \url{https://github.com/zchoi/VCRN}.  
\end{abstract}


\begin{figure}
     \centering
     \includegraphics[width=1.0\linewidth]{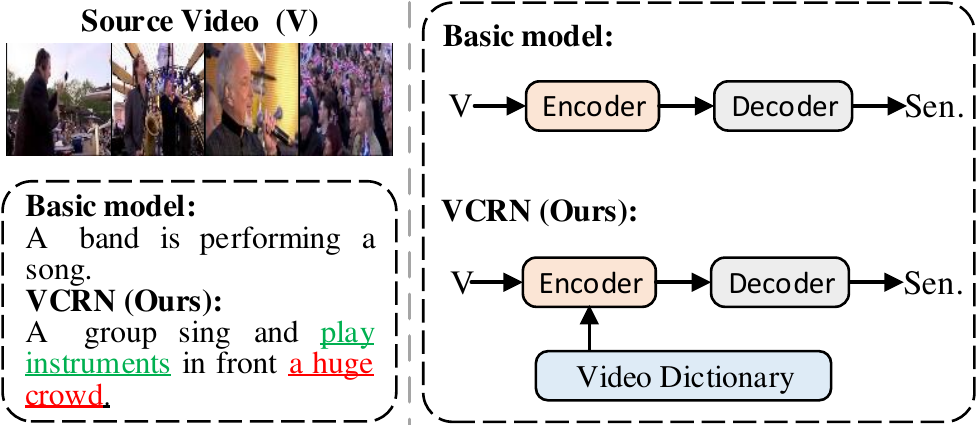}
     \caption{Caption examples from the basic model and VCRN (Ours), respectively. The basic model can only focus on a source video being processed, which is difficult to explore the comprehensive context information about a candidate word, like ``play instruments'' and ``a huge crowd''. In contrast, our proposed VCRN designs a video dictionary to model visual commonsense derived from all videos and captures the correspondence association between the source video and visual commonsense, which yields a more diverse and rich sentence ($Sen.$).}
     \label{fig:intro}
\end{figure}

\section{Introduction}

With the widespread use of mobile phones and computers,
millions of videos are uploaded daily by users to sharing sites such as TikTok, YouTube, and Netflix. Thus, a powerful video captioning method is significantly essential to automatically generate the appropriate descriptions for user-uploaded videos, which can improve the user experience. Besides, there are other broad application scenarios for video captioning, including visually impaired assistance \cite{lee2021goldeye, rane2021image}, online video search \cite{han2021fine, jiang2021learning}, human-computer interaction \cite{dix2000human,das2017learning}, \etc Compared with its twin “image captioning” \cite{RSTNet,9810877} only dealing with static spatial information, video captioning tends to be more challenging since it involves  both consecutive spatial and  temporal representations.

The mainstream approaches for video captioning follow the paradigm of an encoder-decoder framework, where the encoder employs CNNs to analyze and extract useful visual context features from the source video, and the decoder utilizes RNNs to generate the caption sequentially. One effective solution is to learn representative visual features. Toward this goal, existing methods propose a series of attention mechanism by learning the temporal relation between video frames \cite{yao2015describing,yang2017catching}, the spatial relations between objects in every single frame \cite{zhang2020object, tan2021learning}, or spatial-temporal relation using appearance and motion representations \cite{ryu2021semantic,zhao2021multi, gao2022hierarchical}. 

\IEEEpubidadjcol
Although the above methods have achieved remarkable progress, they focus on a source video to exploit spatial-temporal relationships to generate caption via recurrent decoder, which still rely on learning the superficial association contained in the video itself. As an external information, commonsense knowledge is considered a necessary complement to the cross-modal task \cite{zong2021fedcmr,ren2021learning,cai2021ask}, which remains under-explored. For instance, \cite{zhang2020object} designs teacher-recommended learning to take full advantage of the successful external language model (ELM) to integrate rich language knowledge into the captioning model, which only exploits commonsense knowledge in text domain. However, the commonsense knowledge in the video domain~\cite{cao2022visual} is neglected.

Generally, the generated words in the descriptions may occur in multiple video scenes with similar but not identical context information. For instance, the basic model in Fig. \ref{fig:intro}, which is based on the encoder-decoder framework, cannot correspond the information in the source video to the words ``play instruments'' and ``a huge crowd'' accurately because of insufficient visual details. In reality, when comprehending videos, human may also associate
the source video with other videos with similar visual concepts for an analogy to generate more accurate descriptions. Thus, this indicates that video captioning should have the cognitive power of visual commonsense knowledge.

In this paper, we design a novel method for video captioning, called Visual Commonsense-aware Representation Network (VCRN). Since directly modeling the relationship between the source video and other all videos will inevitably increase the computational and time cost, we design a video dictionary to summarize the co-occurrence commonsense knowledge of all videos, so as to explore the association between the source video and visual commonsense knowledge. Our network VCRN comprises the following three major components: 1) \textbf{Video Dictionary construction (VDC)}, which aims to build the commonsense knowledge from a video dataset. Specifically, we employ a K-means algorithm on the video frame representations derived from all videos to yield a video dictionary consisting of a set of cluster centers. And each center is regarded as a visual concept representing one type implicit commonsense knowledge. 2) \textbf{Visual Concept Selection (VCS)}, which is to acquire visual commonsense knowledge related to the source video from the video dictionary. In practice, we adopt a concept-aware multi-head attention to obtain a video-related concept feature by selecting key concept information from the video dictionary guided by the source video. 3) \textbf{Conceptual Integration Generation (CIG)}, which is designed to enhance the caption generation by exploring the relationship between the source video feature and the video-related concept feature. Such a module can provide dynamical control for the propagation of the above two types of features by a gate mechanism. Fig.~\ref{fig:intro} shows that our model can successfully generate fine-grained words ``play instruments'' and ``a huge crowd''  because our method can capture various relevant visual information corresponding to the source video from the video dictionary. To evaluate our proposed method, we conduct extensive experiments and analyses it on the three publicly video captioning benchmarks : MSVD, MSR-VTT, and VATEX. And comprehensive ablation experiments are carried out to prove the effectiveness of our each component.  Besides, to further improve the generalization of our method, our method is successfully applied to video question answering task. Finally, we qualitatively show that our method can contribute to improved captions through case studies.

To summarize, the contributions of this work lie in threefold: 
\begin{itemize}
\item We propose a simple yet effective method, namely a Visual Commonsense-aware Representation Network (VCRN), to explore the effect of visual commonsense information for video captioning, which improves the model's capability of knowledge cognitive.
\item We design a video dictionary, a plug-and-play component, to model visual commonsense and exploit the association between the source video and commonsense via our proposed visual concept selection and conceptual integration generation to yield a more accurate caption.
\item The extensive experimental results demonstrate the benefits of introducing visual commonsense for the video captioning task. The proposed method VCRN achieves state-of-the-art performance on MSVD and VATEX and competitive performance on MSR-VTT. Besides, our approach brings performance gains on video question answering task, further demonstrating the generalization of our method.  
\end{itemize}

\section{Related Works}
\label{related_work}
\subsection{Video Captioning} Video captioning as one of the mainstay in the multi-modal domain, this task has received extensive interest and made rapid development.
With the advent of the encoder-decoder framework, recent researches mainly focus on the sequence-learning based methods for generation process \cite{venugopalan2015sequence,wang2018reconstruction,pan2016hierarchical,DBLP:journals/tnn/SongGGLHS19,yang2021non,tan2021learning}. Technically, these methods employ
an encoder to refine the video representation from a group of fixed video frame features, and then a language-based decoder integrates textual descriptions with the refined video features to learn a modality-aligned representation for caption generation. As one of the precedents that adopt such encoder-decoder structure, \cite{venugopalan2014translating} generates captions by LSTM with mean pooled video representation overall frame features. And \cite{yao2015describing} proposes a temporal attention to dynamically select video frames based on the current decode step. To further align the semantic information between video and language modalities and improve the performance, extensive approaches with elaborate structure \cite{hou2019joint, wang2019controllable, chen2020better, ryu2021semantic, yang2021non} have been proposed. For instance, \cite{ryu2021semantic} encodes a video into semantic groups by aligning frames around the phrases of partially decoded caption and describes the video by exploiting the semantic groups as information units. \cite{chen2019motion} utilizes optical flow to guide the spatial attention, which can capture the pattern of apparent motion between consecutive video frames. 
To improve caption quality, \cite{yang2021non} proposes an alternative paradigm to decompose the captioning procedure into two stages. More recently, there are some methods \cite{zhang2019object, zhang2020object, tan2021learning} have drawn attention to object-level information.
\cite{zhang2019object,bai2021discriminative} adopt a bidirectional temporal graph to capture fine-grained dynamic flow for salient objects in the video. \cite{tan2021learning} performs visual reasoning over both space and time domains then locate region over the video by a spatial-temporal attention.

Unlike these methods, our approach does not introduce extra visual features or pre-trained end-to-end architectures, but mines the underlying semantic knowledge hidden in the datasets, which aims to provide high-level visual concepts for the model reasoning.

\begin{figure*}
    \centering
    \includegraphics[width=\textwidth]{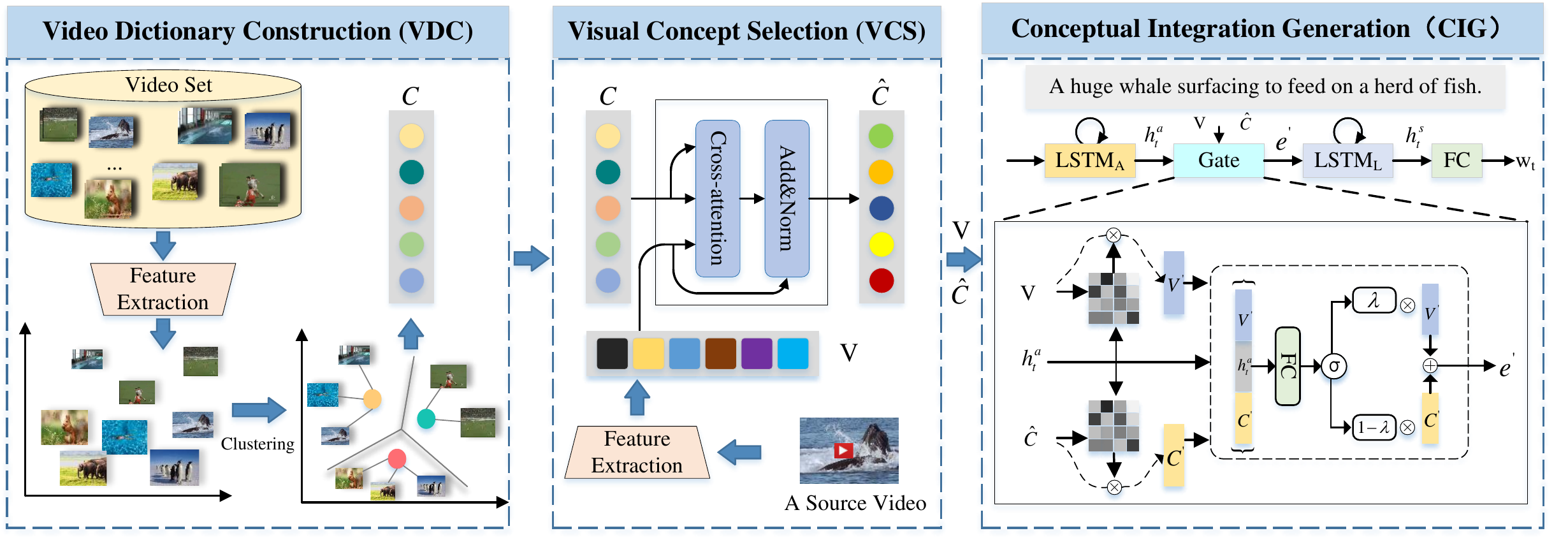}
    \caption{Overview of the proposed Visual Commonsense-aware Representation Network for Video Captioning. It consists of three main components: 1) Video Dictionary Construction, which models the visual commonsense knowledge extracted from all videos, 2) Visual Concept Selection, which aims to learn associations between the current video and the commonsense knowledge, and 3) Conceptual Integration Generation, which is to generate linguistic descriptions by time step. $\textbf{C}$, $\mathbf{\hat C}$ and $\textbf{V}$ denote video dictionary, video-related concept feature and video feature, respectively.}
    \label{fig:framework}
\end{figure*}

\subsection{Knowledge-based Learning} 
To further move towards cognitive understanding of models, many knowledge-based approaches have been proposed \cite{marino2019ok,DBLP:journals/tnn/ZhangLLZLSG21,wen2020multi,salaberria2021image}. In general, most of the existing methods can be categorized into two types. The first one focuses on the structured knowledge base (\eg DBpedia \cite{auer2007dbpedia} and WordNet \cite{miller1995wordnet}) to perform knowledge inference and assist model reasoning. For instance, \cite{wang2017explicit} applies a large-scale knowledge base as visual concepts, \ie ConceptNet \cite{speer2017conceptnet}, for explainable visual question answering (VQA). \cite{zhou2019improving} leverages structured concept graph to improve the performance of image captioning. \cite{wen2020multi} proposes multi-level commonsense knowledge-based learning for visual commonsense reasoning. The other one focuses on the unstructured knowledge base, which explicitly represents knowledge from the linguistic corpus or vision modality. Compared with the structured one, it regularly be acquired through elaborate design such as pre-trained language (LMs) or retrieval model. For instance, \cite{salaberria2021image} 
hypothesizes that a system that relies exclusively on
text will allow LMs to better leverage their implicit
knowledge and then utilize it on visual question answering task.
\cite{zhang2021open} proposes a pluggable retriever to retrieve sentences as prior hints into video captioning model.

Different from previous approaches that exploit consensus knowledge from the external source, our method aims to explore latent association in video set and mine intrinsic commonsense knowledge between videos from inside.

\subsection{Video Question Answering}
Video question answering is another fundamental multi-modal task, which aims to predict an accurate answer according to a video and a corresponding question. The benefit to the success of deep learning, various techniques, \eg attention mechanism~\cite{li2019learnable,HCRN,gao2019structured}, memory network~\cite{gao2018motion,DBLP:conf/acl/GuoZJ0L20}, and graph neural network~\cite{seo2021attend,DBLP:conf/aaai/HuangCZDTG20}, have been proposed to build the relationship between vision and language to answer questions. For instance, \cite{li2019learnable} proposes a temporal attention to focus on the key information through questions as guidance. \cite{gao2018motion} applies a co-memory network to learn the important cues from both motion and appearance and obtain the multi-level contextual facts to infer the answer. \cite{HCRN} introduces a Hierarchical Conditional Relation Network to construct more sophisticated relations across video and question, which obtains diverse modalities and contextual information. \cite{seo2021attend} proposes a Motion-Appearance Synergistic Network to action-oriented cross-modal joint representations between motion and appearance by graph neural network. In this paper, our proposed method is applied to the task of video question answering to verify its effectiveness.

\section{Method}
\label{method}
In this section, we present the proposed video captioning approach based on \textbf{\emph{Visual Commonsense-aware Representation Network (VCRN)} } in detail, which follows the paradigm of the encoder-decoder framework. As shown in Fig.~\ref{fig:framework}, our VCRN consists of three components. \textbf{(1) Video Dictionary Construction (VDC):} For all videos, we first extract motion and appearance features to present video information. Next, we construct a video dictionary to capture and store visual commonsense knowledge in video domain in an unsupervised method. \textbf{(2) Visual Concept Selection (VCS):} Based on the video dictionary, we perform visual concept selection to obtain the video-related concept features via a concept-aware multi-head attention module. \textbf{(3) Conceptual Integration Generation (CIG):} Above two components belong to the encoding stage. While in the decoding stage, the source video features and video-related concept features are fed into the Conceptual Integration Generation to predict descriptions. Especially, a gated controller is applied to distinguish the different contributions of the two above features. In the following subsections, we introduce Visual Dictionary Construction (in Sec.~\ref{VDS}), Visual Concept Selection (in Sec.~\ref{VCS}) and Conceptual Integration Generation (in Sec.~\ref{CIG}), respectively.

\subsection{Video Dictionary Construction}
\label{VDS}
As discussed above, directly operating at a source video to generate description leads to insufficient visual details. A plain idea is to introduce other similar videos to compensate for the deficiency. However, modeling the relationship between the source video and other videos in dataset will inevitably increase the computation burden and time costs of the model. Intuitively, if we implicitly summarize co-occurrence concept information in video domain to represent commonsense knowledge, this operation would become more flexible. Motivated by this, we construct a video dictionary $C$ to obtain intrinsic visual commonsense knowledge in an unsupervised way, containing multiple representative visual concepts.

Concretely, we first employ the 2D-CNN and 3D-CNN to extract appearance feature $\textbf{V}^{a}=\{\textbf{v}_{i}^a\}_{i=1}^L$ and motion feature $\textbf{V}^{m}=\{\textbf{v}_{i}^m\}_{i=1}^L$, respectively, and then concatenate $\textbf{V}^a$ and $\textbf{V}^m$ to sufficiently present a video $\textbf{V}=[\textbf{V}^a; \textbf{V}^m]$, where $L$ denotes the number of frames sampled for this video, $\textbf{v}_{L}$ denotes the visual feature of $L$-th frame, and [;] means the concatenate operation. 
Based on the above process, we extract the features of all video $\textbf{V}_{all}$ in the dataset. Afterward, we utilize the K-means algorithm to apply upon these video features $\textbf{V}_{all}$ to obtain $M$ cluster centers, which is denoted as $\textbf{C} = \{\textbf{c}_1,...,\textbf{c}_M\}$, where $\textbf{c}_j$ is regarded as the $j$-th visual concept representation. Thus, we define the final $\textbf{C}$ as a video dictionary, which will be used for assisting the original video in obtaining additional visual commonsense knowledge.


\subsection{Visual Concept Selection} \label{VCS}

The goal of visual concept selection (VCS) is to learn the key commonsense knowledge from a Video Dictionary to obtain a video-related concept feature $\mathbf{\hat C}$. The main architecture of VCS applies Concept-aware Cross Attention module (C-MCA). As shown in Fig.~\ref{fig:framework}, we first use different fully-connected (FC) layers to map $\textbf{V}=\{\textbf{v}_i\}_{i=1}^{L}$ to queries ($\textbf{Q}_v$) and $\textbf{C}=\{\textbf{c}_j\}_{j=1}^{M}$ to keys ($\textbf{K}_
c$) and values ($\textbf{V}_c$). The similarity matrix $\textbf{S}$ between the video feature $\textbf{V}$ and all concepts in the video dictionary $\textbf{C}$ is calculated by a scaled dot-product attention function:
\begin{equation} \label{eq1}
    \textbf{S}= Similarity({\textbf{Q}_v},{\textbf{K}_c}) = \textsc{softmax} (\frac{{{\textbf{Q}_v}\textbf{K}_c^T}}{{\sqrt d }}).
\end{equation}
Here, the similarity $\textbf{S}_{(i,j)}$ indicates the degree that the $j$-th concept feature $\textbf{c}_j$ should attend to the $i$-th video feature $\textbf{v}_i$. To focus on multiple semantically related visual concepts simultaneously, we adapt Multi-Head Attention (MHA) to re-calculate the similarity S in Eq.~\ref{eq1}.

In MHA, several projection matrices of queries $\textbf{Q}_v$, values $\textbf{K}_c$, and keys $\textbf{V}_c$ are used for different heads, and these matrices is mapped into different sub-spaces. Subsequently, the similarity $\textbf{S}^{(h)}$ of head $h$ is calculated by Eq.~\ref{eq1} to aggregate multiple semantic information between query $\textbf{Q}_{v}^{(h)}$ and key $\textbf{K}_{c}^{(h)}$. All similarity head are concatenated together and fused with the learnable projection $\textbf{W}^O$: 
\begin{equation}
    \begin{array}{l}
\textbf{C}_s^{(h)} = \textsc{Dropout}({\textbf{S}^{(h)}}\textbf{V}_c^{(h)}), \  \textsc{for}\ h = 1,2,...,H,\\
{\textbf{C}_s} = [\textbf{C}_s^{(1)};\textbf{C}_s^{(2)};...;\textbf{C}_s^{(H)}]{\textbf{W}^O},
\end{array}
\end{equation}
where H is the number of heads and $\textbf{C}_{s}^{(h)}$ is the output of the $h$-th head. Finally, $\textbf{C}_{s}$ is normalized via Layer Normalization and added to a source video feature to produce a video-related concept feature $\textbf{C}_{t}$:
\begin{equation}
    \begin{array}{c}
    \mathbf{C}_{t} = \textbf{V} + \textsc{LayerNorm}(\textbf{C}_s).
\end{array}
\end{equation}

We stack $N$ C-MCA blocks to obtain a more refined video-related concept feature, and take the output of the last C-MCA block as a final video-related concept feature $\mathbf{\hat C}$.

\subsection{Conceptual Integration Generation} \label{CIG}
At the decoding stage, we design a novel conceptual integration generation (CIG) to generate captions based on source video feature and video-related concept video feature. The CIG is composed of three parts: an Attention-LSTM, a Gated Controller, and a Language-LSTM. We describe the proposed generator in detail as follows:

\textbf{Attention-LSTM.} At the $t$-th time step, the Attention-LSTM ($LSTM_{A}$) aims to obtain the semantics of the current state ${\rm{\textbf{h}}}_t^{\rm{a}}$ according to the previous hidden state $\textbf{h}_{t-1}^l$ of the Language-LSTM, concatenated with global video feature $\mathbf{\bar v}$ and the previous word $w_{t-1}$:
\begin{equation}
\begin{array}{l}
{\bf{h}}_t^{\rm{a}} = LSTM_A([{\bf{h}}_{t - 1}^l;\mathbf{\bar v}; {{\bf{W}}_e}{{{w}}_{t - 1}}],{\bf{h}}_{t - 1}^a),\\
\mathbf{\bar v} = \frac{1}{L}\sum {{\textbf{v}_l}}, 
\end{array}
\end{equation}
where $[;]$ means the the operation of concatenation, and $\textbf{W}_e$ denotes the word embedding matrix.

\textbf{Gated Controller.} The designed gated controller is adopted for the aggregation of video representations $\textbf{V}$ and $\mathbf{\hat C}$ according to the current hidden state $\textbf{h}_{t}^a$ of the attention-LSTM, it enables which information flows (\ie $\textbf{V}$ and $\mathbf{\hat C}$) should play a more important role in the language-LSTM. Specifically, we first apply multiplicative attention-mechanism to aggregate video feature $\textbf{V}$ with the current hidden state $\textbf{h}_{t}^a$ at frame-level to obtain the attended video feature $\textbf{V}^{'}$:
\begin{equation}
	\begin{array}{l}
		{\textbf{V}^{'}} = \sum\limits_{i = 1}^L {{\alpha _i}{\textbf{V}}_i}, \\
		{\alpha _i} = \textsc{softmax} ({\textbf{W}_1}\tanh ({\textbf{W}_2}{\textbf{V}}_i \oplus {\textbf{W}_3}\textbf{h}_t^a)),
	\end{array}
\end{equation}
where $\oplus$ is element-wise addition, $\textbf{W}_{*}$ is the learnable matrices. $\textbf{V}_i$ means the $i$-th frame-level vector in the video feature $\textbf{V}$. To simplify the process of $\textbf{V}^{'}$ extraction, we formulate it as:
\begin{equation}
	\begin{array}{c}
		\textbf{V}^{'}=ATT_{V}(\textbf{V},\textbf{h}_{t}^{a}).
	\end{array}
\end{equation}
Similar to the operation of $ATT_{V}$, we integrate current
hidden state $\textbf{h}_{t}^a$ with video-related concept feature to produce the attended concept feature $\textbf{C}^{'}$. The extraction module of $\textbf{C}^{'}$ is defined as:
\begin{equation}
	\begin{array}{c}
		\textbf{C}^{'}=ATT_{C}(\mathbf{\hat C},\textbf{h}_{t}^{a}).
	\end{array}
\end{equation}

As illustrated in Fig. \ref{fig:framework}, the context gate $\lambda$ controls the propagation of the information of  $\textbf{V}^{'}$ and $\textbf{C}^{'}$ to Language-LSTM. In practice, the value of $\lambda$ is based on $\textbf{V}^{'}$, $\textbf{C}^{'}$ and $\textbf{h}_{t}^a$ via a nonlinear layer:
\begin{equation}
	{\lambda} = \sigma(\textbf{W}_{\lambda} \cdot [\textbf{V}^{'}; \textbf{C}^{'}; \textbf{h}_{t}^a]),
\end{equation}
where $\textbf{W}_{\lambda}$ is a learnable parameter and $\sigma(\cdot)$ denotes the sigmoid function. Next, we utilize this gated controller in a bilateral scheme, where $\lambda$ determines the flow of $\textbf{V}^{'}$ and the complementary part $1-\lambda$ governs the amount of $\textbf{C}^{'}$, to get conceptual integrated video feature $\textbf{e}^{'}$:
\begin{equation}
	{\textbf{e}^{'}} = \lambda \odot f(\textbf{V}^{'})+ (1-\lambda) \odot f(\textbf{C}^{'}),
	\label{11}
\end{equation}
where $\odot$ is Hadamard product, $f(\cdot)$ can be represented as fully-connected layer or identity mapping.

\textbf{Language-LSTM.} The language-LSTM feeds the hidden conceptual integrated video feature $\textbf{e}^{'}$ to generate the current hidden state $\textbf{h}_{t}^l$.  The logits distribution of the caption model ${\textbf{p}_t}$ is acquired via a single linear function and the sofrmax operation at the decoding step $t$:
\begin{equation}
	\begin{array}{c}
		{\textbf{h}_{t}^l} = LSTM_{L}([\textbf{h}_{t}^a;\textbf{e}^{'}],\textbf{h}_{t-1}^l), \\
		{\textbf{p}_t} = \textsc{softmax}(\textbf{W}_v\textbf{h}_{t}^l+\textbf{b}_v),
	\end{array}
\end{equation}
where $\textbf{p}_t$ is a vector of the vocabulary size and $\textbf{W}_v$ and $\textbf{b}_v$ are learnable parameters.

Following the standard objective of video captioning, we adopt Cross-Entropy loss to optimize our model:
\begin{equation}
	{L_{CE}} =  - \sum\limits_{t = 1}^T {\log ({p_\theta }(w_t^*|w_{ < t}^*))},
\end{equation}
where $w_{1:T}^{*}$ is the target ground-truth sequences, and $\theta$ is the parameters of our captioning model.

\begin{table*}[]
\centering
\caption{Performance comparisons on MSVD and MSR-VTT datasets. All the approaches are divided into two categories. Top rows use three vision features, including appearance feature, motion feature and object feature, to train the model while bottom rows only utilize appearance feature and motion feature to train the model.}
\label{tab:tab1}
\resizebox{1\textwidth}{!}{%
\begin{tabular}{c|ccc|cccc|cccc}
\toprule
\multirow{2}{*}{Models} & \multicolumn{3}{c|}{Features}              & \multicolumn{4}{c|}{MSVD}                                      & \multicolumn{4}{c}{MSR-VTT}                                   \\
                        & Appearance   & Motion       & Object       & BLEU-4           & METEOR             & ROUGE-L             & CIDEr              & BLEU-4           & METEOR             & ROUGE-L             & CIDEr            \\ \midrule
OA-BTG \cite{zhang2019object}                  & $\checkmark$ & $\times$     & $\checkmark$ & 56.9          & 36.2          & -             & 90.6           & 41.4          & 28.2          & -             & 46.9          \\
MGSA \cite{chen2019motion}                   & $\checkmark$ & $\checkmark$ & $\checkmark$ & 53.4          & 35.0          & -             & 86.7           & 42.4          & 27.6          & -             & 47.5          \\
STG \cite{pan2020spatio}                    & $\checkmark$ & $\checkmark$ & $\checkmark$ & 52.2          & 36.9          & 73.9          & 93.0           & 40.5          & 28.3          & 60.9          & 47.1          \\
SAAT \cite{zheng2020syntax}                   & $\checkmark$ & $\checkmark$ & $\checkmark$ & 46.5          & 33.5          & 69.4          & 81.0           & 40.5          & 28.2          & 60.9          & 49.1          \\
RMN \cite{tan2020learning}                & $\checkmark$ & $\checkmark$ & $\checkmark$ & 54.6          & 36.5          & 73.4          & 94.4           & 42.5 & 28.4          & 61.6          & 49.6          \\
MGRMP \cite{chen2021motion}                  & $\checkmark$ & $\checkmark$ & $\checkmark$ & 53.2          & 35.4          & 73.5          & 90.7           & 42.1          & 28.8 & 61.4          & 50.1          \\ 
ORG-TRL \cite{zhang2020object}         & $\checkmark$ & $\checkmark$ & $\checkmark$     & 54.3          & 36.4          & 73.9          & 95.2           & \textbf{43.6}         & \textbf{28.8}          & \textbf{62.1}   & \textbf{50.9}    \\  \midrule  \midrule
MARN \cite{pei2019memory}                   & $\checkmark$ & $\checkmark$ & $\times$     & 48.6          & 35.1          & 71.9          & 92.2           & 40.4          & 28.1          & 60.7          & 47.1          \\
M3 \cite{wang2018m3}                     & $\checkmark$ & $\checkmark$ & $\times$     & 52.8          & 33.3          & -             & -              & 38.1          & 26.6          & -             & -             \\
POS-CG \cite{wang2019controllable}                 & $\checkmark$ & $\checkmark$ & $\times$     & 52.5          & 34.1          & 71.3          & 88.7           & 42.0          & 28.2          & 61.6 & 48.7          \\
MDT \cite{zhao2021multi}                     & $\checkmark$ & $\checkmark$ & $\times$     & 49.0          & 35.3          & 72.2          & 92.5           & 40.2          & 28.2          & 61.1          & 47.3          \\
SGN \cite{ryu2021semantic}                     & $\checkmark$ & $\checkmark$ & $\times$     & 52.8          & 35.5          & 72.9          & 94.3           & 40.8          & 28.3          & 60.8          & 49.5          \\ 
HRNAT \cite{gao2022hierarchical}                     & $\checkmark$ & $\checkmark$ & $\times$     & 55.7          & 36.8          & 74.1          & 98.1           & 42.1          & 28.0         & 61.6         & 48.2       \\ 
VCRN (ours)              & $\checkmark$ & $\checkmark$ & $\times$     & \textbf{59.1} & \textbf{37.4} & \textbf{74.6} & \textbf{100.8} & 41.5          & 28.1          & 61.2          & 50.2 \\ \bottomrule
\end{tabular}%
}
\end{table*}

\section{Experiments}
\label{experiments}


\subsection{Datasets and Metrics}
\subsubsection{Datasets} Following the previous works \cite{zhang2020object}, we evaluate our method VCRN on three publicly available datasets: MSVD, MSR-VTT and VATEX, for video captioning.

\textbf{MSVD} \cite{chen2011collecting} is a collection of 1,970 short clip videos downloaded from YouTube website. Each clip has 35 captions annotated by humans. To be consistent with the previous works, we use standard splits, namely 1,200 clips for training, 100 clips for validation, and 670 clips for testing.

\textbf{MSR-VTT} \cite{xu2016msr} consists 10,000 open domain videos from YouTube with 20 human descriptions for each video clip. We follow the standard split with 6,573 videos for training, 497 videos for validation, and the remaining 2,990 for testing.

\textbf{VATEX} \cite{wang2019vatex} is a recently released large-scale multilingual video description dataset, which reuses the video source from Kinetics-600. It contains over 41,250 videos, where each video clip is annotated with 10 English and Chinese descriptions respectively. In this paper, we only utilize English captions for our experiments. According to the official split, the dataset is divided into 25,991 for training, 3,000 for validation, and 6,000 for public testing.

Besides, to verify the generalization of our method, we also conduct experiments on two video question answering (VideoQA) datasets: MSVD-QA and MSRVTT-QA.

\textbf{MSVD-QA} \cite{xu2017video} is derived from the existing MSVD dataset with the same video data, containing 1,970 short clips and 50,505 question-answer pairs. These question-answer pairs are split into five types according to question purpose: what, where, when, how, and who.

\textbf{MSRVTT-QA} \cite{xu2016msr} is composed of 10K videos from MSR-VTT dataset and 243K annotated question-answer pairs, where the questions are also of five types. Compared to MSVD-QA, the video length of MSRVTT-QA is much longer, roughly around 10-30 seconds with more complex scenes.

\subsubsection{Evaluation Metrics} For captioning task, we employ the standard captioning evaluation metrics, including BLEU-4 \cite{papineni2002bleu}, METEOR \cite{denkowski2014meteor}, CIDEr \cite{vedantam2015cider}, and ROUGE-L \cite{lin2004rouge}, to evaluate our method. 
For VideoQA, the accuracy is adopted as the evaluation metric.


\subsection{Implementation Details}
\subsubsection{Feature Extraction} For the visual features, we use ResNet \cite{he2016deep} as 2D CNN and ResNeXt \cite{xie2017aggregated} as 3D CNN from the MXNet library \cite{chen2015mxnet} to extract appearance feature and motion feature, respectively. The above features are extracted from 26 keyframes of videos by equally interval sampling.

For caption, we remove punctuation, convert all words to lower case and keep the words that occur more than 2 times for MSR-VTT and MSVD (5 for VATEX) to a word vocabulary. Descriptions longer than 26 words (30 for VATEX) will be truncated for the convenience of training. Besides, we add three special tokens (``$<bos>$'',  ``$<eos>$'' and ``$<pad>$'') to the word vocabulary. GloVe \cite{pennington2014glove} is utilized to initialize the word embedding.

\subsubsection{Training Details} We adopt Adam \cite{kingma2014adam} optimizer with the learning rate of 1e-4 to train our model. We choose hyperparameter M=1,000 as the number of clustered centers, N=1 for MSVD, and N=3 for MSR-VTT and VATEX. The batch size is set to 64 for all datasets. The hidden size of the LSTM is 512, 1024, and 1024 for MSVD, MSR-VTT, and VATEX respectively. During the testing phase, we set beam size with 5 for MSVD, 2 for MSR-VTT and VATEX. All experiments will be completed after 20 epochs. We implement our VCRN method by PyTorch and run on one NVIDIA V100 GPU.

\begin{table}[]
\centering
\caption{Performance comparisons on VATEX testing set. Note that the ORG-TRL \cite{zhang2020object} employs three features (\ie appearance feature, motion feature, and object feature), and exploits linguistic knowledge, while the rest methods including ours only adopt appearance feature and motion feature. B@4, M, R, C indicate BLEU-4, METEOR, ROUGE-L, CIDEr, respectively.
}
\label{tab:ab4}
\resizebox{0.85\linewidth}{!}{%
\begin{tabular}{ccccc}
\toprule
Model                              & B@4           & M             & R                  & C             \\ \midrule 
ORG-TRL \cite{zhang2020object}      & 32.1          & 22.2          & 48.9 & 49.7          \\ \midrule
Shared Base \cite{wang2019vatex}    & 28.1          & 21.6          & 46.9               & 44.3          \\
Shared Enc \cite{wang2019vatex}     & 28.4          & 21.7          & 47.0               & 45.1          \\
Shared Enc-Dec \cite{wang2019vatex} & 27.9          & 21.6          & 46.8               & 44.2          \\
HRNAT \cite{gao2022hierarchical}    & 32.1          & 21.9          & 48.4               & 48.5          \\ 
Ours                               & \textbf{32.4} & \textbf{22.4} & \textbf{48.9}      & \textbf{49.9} \\ \bottomrule
\end{tabular}%
}
\end{table}

\subsection{Performance Comparisons}
\textbf{Compared Methods.} In this section, we compare our method VCRN with the state-of-the-art approaches on MSVD, MSR-VTT, and VATEX datasets. These state-of-the-art approaches can be divided into two categories: \romannumeral1) The first category employs appearance feature, motion feature and object feature to train their model, including OA-BTG \cite{zhang2019object}, MGSA \cite{chen2019motion}, STG \cite{pan2020spatio}, SAAT \cite{zheng2020syntax}, RMN \cite{tan2020learning}, MGPMP \cite{chen2021motion}, ORG-TRL \cite{zhang2020object}; and \romannumeral2) The second category only utilizes appearance feature and motion feature without help of object feature, including MARN \cite{pei2019memory}, M3 \cite{wang2018m3}, POS-CG \cite{wang2019controllable}, MDT \cite{zhao2021multi}, SGN \cite{ryu2021semantic} and HRNAT \cite{gao2022hierarchical}, Shared Base \cite{wang2019vatex}, Shared Rnc \cite{wang2019vatex}, and Shared Enc-Dec \cite{wang2019vatex}. Here, our model belongs to the second category.

\textbf{Comparisons on MSVD.} The results of comparison on MSVD are reported on Tab.~\ref{tab:tab1}. We can find that our VCRN model exceeds all previous models in all metrics (BLEU-4, METEOR, ROUGE-L and CIDEr). Compared
with the second category methods, our method outperforms the best counterpart HRNAT, especially with an increase of 3.4\% and 2.7\% in terms of BLEU-4 and CIDEr, respectively. Compared with the first category methods, our model can still significantly outperform them by a large margin and in particularly increases BLEU-4 and CIDEr by 4.8\% and 5.6\%, respectively. It clearly demonstrates the effectiveness of
our method.


\textbf{Comparisons on MSR-VTT.} 
Tab.~\ref{tab:tab1} also shows the results of comparison on MSR-VTT dataset. We can see that our model maintains relatively comparable performance compared to the existing methods. Although the improvement in MSR-VTT dataset is not obvious as in MSVD dataset, our method gets second place with CIDEr of 50.2\%. Specifically, our method achieves better performance in CIDEr, compared to the second category methods, in particularly obtaining 2.0\% relative gains. Compared to the best counterpart ORG-TRL belonging to the first category, our method is only slightly degraded in performance. The possible reason may be that the best method ORG-TRL belonging to the first category use additional language models, in addition to introducing object features, which is helpful for reasoning on MSR-VTT dataset than MSVD dataset.


\textbf{Comparisons on VATEX.} To further verify the robustness of our method, we provide quantitative results on VATEX dataset in Tab.~\ref{tab:ab4}. From the table, we can observe that our method shows better superiority over all compared methods in all metrics. In particular, in terms of CIDEr, our method brings an increase of 0.2\% and 1.4\% compared to the second (ORG-TRL) and third (HRNAT) methods, respectively. These results well demonstrate the effectiveness of our method.


\begin{table}[]
\centering
\caption{Ablation studies of the proposed Visual Concept Selection (VCS) and Conceptual Integration Generation (CIG).
B@4, M, R, C indicate BLEU-4, METEOR, ROUGE-L, CIDEr, respectively.
}
\label{tab:abl1}
\resizebox{\linewidth}{!}{%
\begin{tabular}{ccccccccc}
\toprule
\multirow{2}{*}{Methods} & \multicolumn{4}{c}{MSVD}                                       & \multicolumn{4}{c}{MSR-VTT}                                   \\ \cmidrule(l){2-9}
                         & B@4           & M             & R             & C              & B@4           & M             & R             & C             \\ \midrule
Baseline (B)              & 57.9          & 36.7          & 74.3          & 96.4           & 40.8          & 27.7          & 60.6          & 48.0          \\
B + VCS              & \textbf{59.5} & 36.8          & 74.2          & 98.1           & 41.2 & 27.9          & 61.1          & 49.6          \\
B + VCS + CIG    & 59.1          & \textbf{37.4} & \textbf{74.6} & \textbf{100.8} & \textbf{41.5}          & \textbf{28.1} & \textbf{61.2} & \textbf{50.2} \\ \bottomrule
\end{tabular}%
}
\end{table}


\begin{figure}[!t]
\centering
\subfloat[\small{MSVD}]{\includegraphics[width=1.5in]{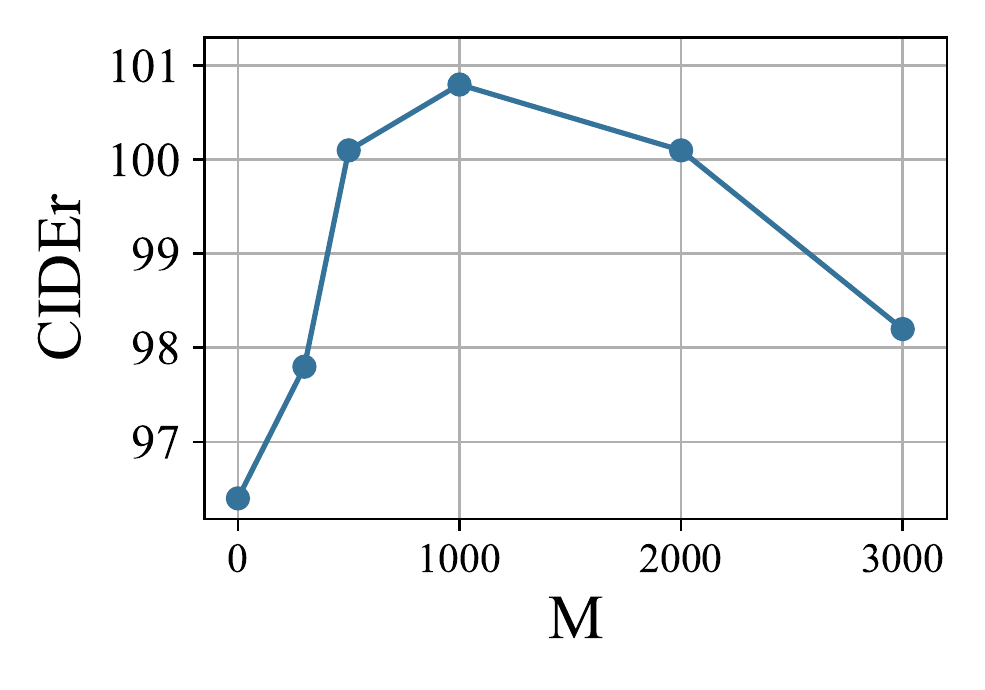}%
\label{fig:a}}
\hfil
\subfloat[\small{MSR-VTT}]{\includegraphics[width=1.5in]{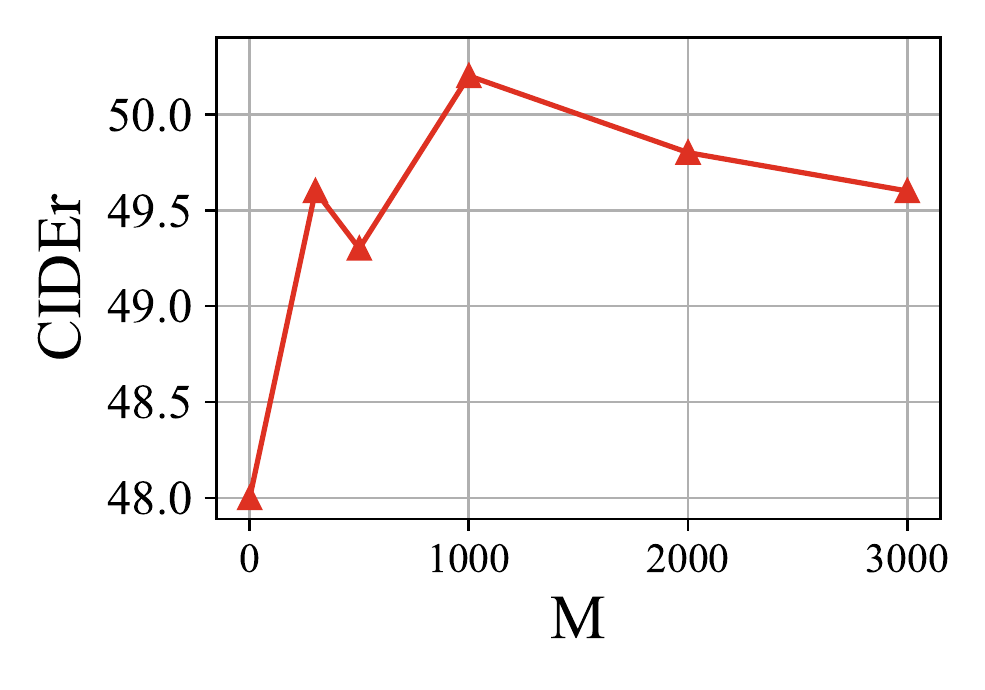}%
\label{fig:b}}
\caption{Ablation studies of the number of clustered centers M in visual dictionary on CIDEr metric.}
\label{fig:N_and_x}  
\end{figure}

\begin{figure}[!t]
\centering
\subfloat[\small{MSVD}]{\includegraphics[width=1.5in]{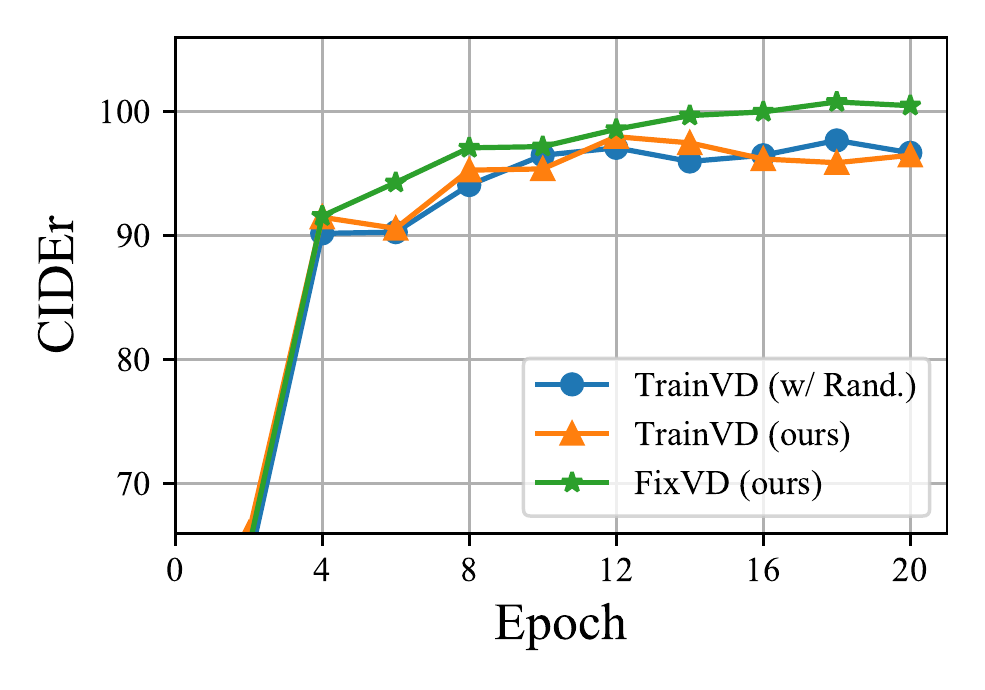}
\label{fig:a}}
\hfil
\subfloat[\small{MSR-VTT}]{\includegraphics[width=1.5in]{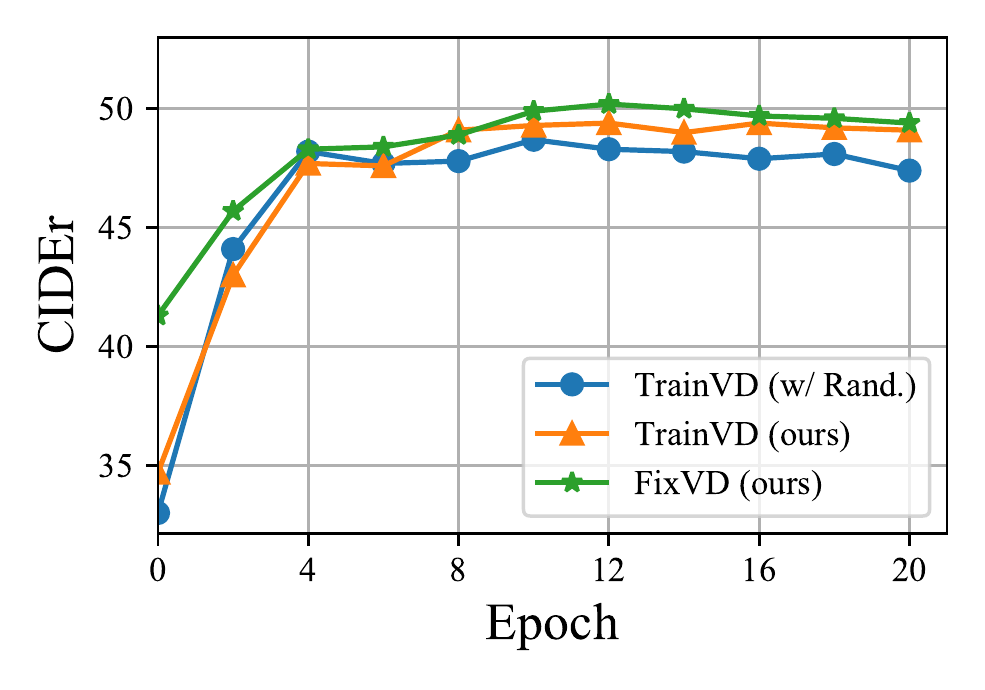}
\label{fig:b}}
\caption{Ablation studies of different training strategies for video dictionary during training on CIDEr metric. ``TrainVD'' and ``FixedVD'' denote the jointly trained video dictionary and fixed video dictionary separately. ``Rand.'' and ``ours'' mean the video dictionary is initialized by random parameters and our constructed method, respectively.}
\label{fig:joint} 
\end{figure}

\subsection{Ablation Study}
In this section, we elaborate on a series of ablation studies in the following Q\&As to better prove the validity of our model. All experimental results are conducted on MSVD and MSR-VTT.

\textbf{Does each component of VCRN affect the results?}
We evaluate the effectiveness of each component by taking successively visual concept selection (VCS) and conceptual integration generation (CIG) into the baseline, where the baseline only adopts appearance feature and motion feature and is based on vanilla encoder-decoder with temporal attention. The results are shown in Tab.~\ref{tab:abl1}. Overall, all the proposed components contribute significantly to the overall performance. Specifically,
the baseline model first performs the worst. By integrating the VCS into the baseline, the performance obtains larger improvement, particularly increased by 1.7\% and 1.6\% in terms of CIDEr on MSVD and MSR-VTT, respectively. It reveals the importance of visual commonsense knowledge, which provides additional visual information to help model reasoning. Then, the CIG is added to the model B+VCS, which in turn further enhances the performance, indicating our CIG can effectively integrate the original video feature and video-related concept feature from VCS.

\begin{table}[]
\centering
\caption{Ablation studies of using different qualities of video dictionary. B@4, M, R, C indicate BLEU-4, METEOR, ROUGE-L, CIDEr, respectively.}
\label{tab:traintest}
\resizebox{\linewidth}{!}{%
\begin{tabular}{cccccccccc}
\toprule
\multirow{2}{*}{Concept Source} & \multirow{2}{*}{Size} & \multicolumn{4}{c}{MSVD}   & \multicolumn{4}{c}{MSR-VTT} \\ \cmidrule(l){3-10} 
                                &                       & B@4  & M    & R    & C     & B@4   & M     & R    & C    \\ \midrule
Train Set                          & +1\%                   & 57.2 & 36.7 & 74.0 & 96.6  & 40.9  & 27.8  & 60.7 & 48.6 \\
Train Set                           & +10\%                  & 57.1 & 36.7 & 74.0 & 97.4  & 41.1  & 28.0  & 61.1 & 49.4 \\
Train Set                           & +50\%                  & 58.8 & 37.1 & 74.3 & 99.8  & 41.3  & 28.1  & 61.2 & 49.6 \\
Train Set                           & +100\%                 & 59.1 & 37.4 & 74.6 & 100.8 & 41.5  & 28.1  & 61.2 & 50.2 \\
Test Set                           & +100\%                 & 59.0 & 37.0 & 74.3 & 98.7  & 41.2  & 28.1  & 61.2 & 49.6 \\ \bottomrule
\end{tabular}%
}
\end{table}

\begin{table}[]
\centering
\caption{Ablation studies of cross-dataset video dictionary. B@4, M, R, C indicate BLEU-4, METEOR, ROUGE-L, CIDEr, respectively.}
\label{tab:across}
\resizebox{\linewidth}{!}{%
\begin{tabular}{ccccccccc}
\toprule
\multirow{2}{*}{Concept Source} & \multicolumn{4}{c}{MSVD}   & \multicolumn{4}{c}{MSR-VTT} \\ \cmidrule(l){2-9} 
                                & B@4  & M    & R    & C     & B@4   & M     & R    & C    \\ \midrule
MSVD                            & 59.1 & 37.4 & 74.6 & 100.8 & 40.4  & 27.7  & 60.5 & 48.9 \\
MSR-VTT                         & 57.9 & 37.5 & 74.4 & 97.8  & 41.5  & 28.1  & 61.2 & 50.2 \\ \bottomrule
\end{tabular}%
}
\end{table}

\textbf{Does the number of clustered centers $M$ in visual dictionary affect the results?} We exploit how the number of clustered centers in visual dictionary affects the performance of our VCRN. In experiments, we select different clustered centers for retraining our VCRN, where $M \in [0, 300, 500, 1000, 2000, 3000]$. Here $M=0$ is treated as baseline without including VCS and CIG, and we pick CIDEr as the metric of caption performance as it reflects the generation relevant to video content.
Fig.~\ref{fig:N_and_x} shows the experimental results. We can see that the performance is best when M is set to 1,000. When the number of clusters is greater than 1,000 or less than 1,000, there exists degradation of model performance. An intuitive explanation is that too many or too few clustered centers can lead to the introduction of redundant or insufficient visual commonsense information, respectively. Thus, we set $M=1,000$ in the final model.


\textbf{Which is better, fixed or jointly trained visual dictionary?}
Fig.~\ref{fig:joint} displays the real-time test results of fixed visual dictionary and jointly trained visual dictionary during training, where we also choose CIDEr as the main metric, which have three settings: \romannumeral1) TrainVD (w/ Rand.): jointly trained visual dictionary initialized by random parameters; \romannumeral2) TrainVD (ours): jointly trained visual dictionary initialized by our proposed video dictionary construction method; and \romannumeral3) FixedVD (ours): fixed visual dictionary initialized by our proposed video dictionary construction method. From the Fig~\ref{fig:joint}, it is observed that FixdVD (ours) is better than  TrainVD (ours) and TrainVD (w/ Rand.). This may be because the fixed video dictionary can retain more original visual commonsense knowledge that is more helpful for generation compared to jointly trained video dictionary.



\begin{table}[]
\centering
\caption{Ablation study of different control strategies in CIG. B@4, M, R, C indicate BLEU-4, METEOR, ROUGE-L, CIDEr, respectively.}
\label{tab:abl2}
\resizebox{\linewidth}{!}{%
\begin{tabular}{ccccccccc}
\toprule
\multirow{2}{*}{Fusion strategy} & \multicolumn{4}{c}{MSVD}                                       & \multicolumn{3}{c}{MSR-VTT}                                   \\ \cmidrule(l){2-9} 
                                 & B@4           & M             & R             & C              & B@4           & M             & R             & C             \\ \midrule
ADD                         & \textbf{59.5} & 36.8          & 74.2          & 98.1           & 41.2 & 27.9          & 61.1          & 49.6          \\
MLP                              & 57.5          & 37.0          & 74.3          & 96.4           & 40.8          & 28.0          & 61.1          & 49.2         \\
MHA                              & 55.9          & 36.6          & 73.6          & 99.2           & 40.2          & 27.9          & 60.7          & 48.7          \\ \midrule
GATE(ours)                       & 59.1          & \textbf{37.4} & \textbf{74.6} & \textbf{100.8} & \textbf{41.5}          & \textbf{28.1} & \textbf{61.2} & \textbf{50.2} \\ \bottomrule
\end{tabular}%
}
\end{table}

\begin{table}[]
\centering
\caption{Generalization of our proposed method on VideoQA task. All values are reported as accuracy (\%).}
\label{tab:ab5}
\resizebox{0.8\linewidth}{!}{
\begin{tabular}{lc|cc}
\toprule
Models                & VDC & MSVD-QA       & MSRVTT-QA     \\ \midrule
\multirow{2}{*}{HME \cite{fan2019heterogeneous}}   & \XSolidBrush   & 33.4          & 32.8          \\
                      & \CheckmarkBold   & \textbf{34.0} & \textbf{33.3} \\ \midrule
\multirow{2}{*}{MASN \cite{seo2021attend}}  & \XSolidBrush   & 36.3          & 34.8          \\
                      & \CheckmarkBold   & \textbf{36.8} & \textbf{35.3} \\ \midrule
\multirow{2}{*}{HCRN \cite{HCRN}}  & \XSolidBrush   & 36.2          & 35.2          \\
                      & \CheckmarkBold   & \textbf{37.4} & \textbf{35.9} \\ \bottomrule
\end{tabular}
}
\end{table}

\begin{figure*}
    \centering
    \includegraphics[width=\textwidth]{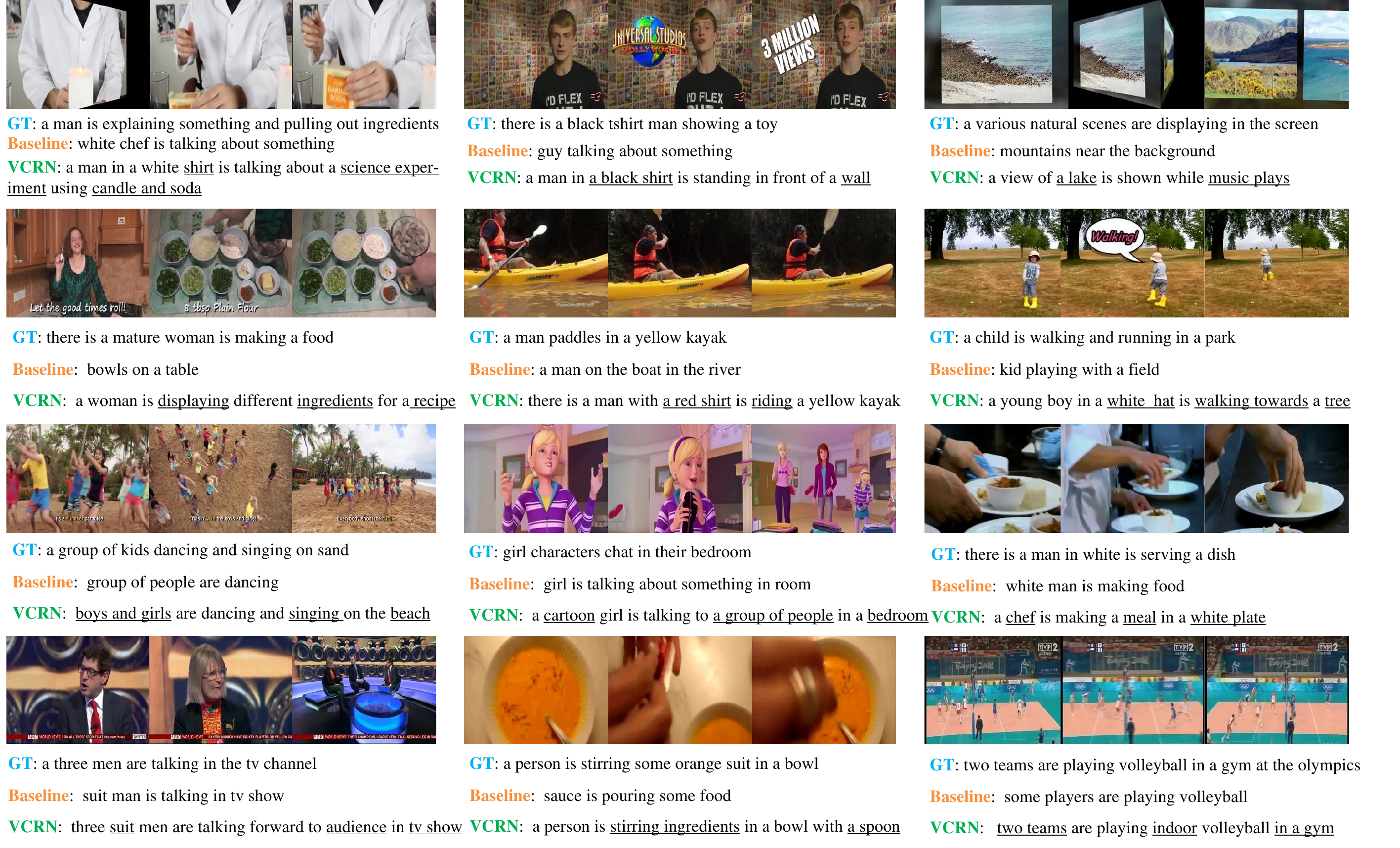}
    \caption{Visualization of the baseline model and our proposed VCRN on MSR-VTT. Each example consists of a raw video, a ground-truth description and the generated descriptions by baseline and ours.}
    \label{fig:vis}
\end{figure*}

\begin{figure}
    \centering
    \includegraphics[width=\linewidth]{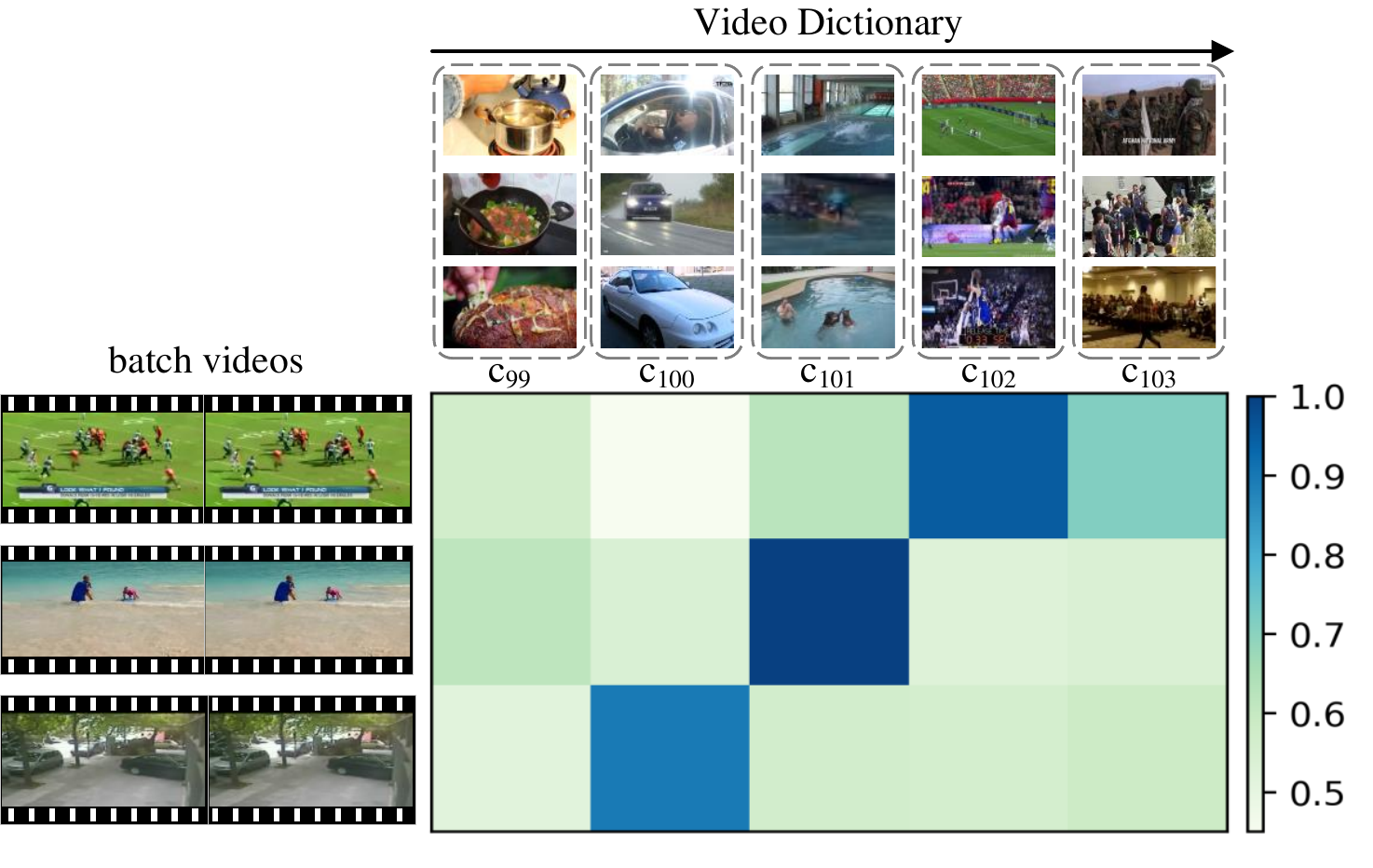}
    \caption{Visualization of the attention map between videos and a partial video dictionary in a batch. Each concept of the video dictionary is represented by a number of video frames.}
    \label{fig:heatmap}
\end{figure}

\textbf{Does the quality of the video dictionary affect the results?} We analyze the effect of randomly selecting different proportions of video in train set to simulate different quality video dictionaries. The experimental results are shown in Tab.~\ref{tab:traintest}. From the table lines 1 to 4, it illustrates that higher quality video dictionaries facilitate the generation of fine-grained descriptions. It can be explained by the fact that higher quality video dictionaries contain more visual commonsense knowledge and provide more hints related to video content for generation. Furthermore, 
we construct the video dictionary using the video data from the test set rather than the training set. It can be seen in Tab.~\ref{tab:traintest} line 5 that the performance drops slightly, which shows that our model has a strong generalization capability.


\textbf{Does cross-dataset video dictionary affect the results?}
We conduct this experiment by using different video dictionary constructed from other video datasets to demonstrate the generalization of the model. In our experiments, we use the video dictionary from the MSR-VTT dataset for training and testing on the MSVD dataset, and vice versa. As shown in Tab.\ref{tab:across}, the model can maintain competitive performance with only a slightly drop when using other video dictionary. It proves that our model has a strong learning ability, which can be extended by changing different video dictionary that are not strongly correlated even with the test data..

\textbf{Does the control strategies in CIG affect the results?}
We compare the effect of different control strategies in CIG, including 1) ADD: adopting an element-wise addition to aggregate $V^{'}$ and $C^{'}$, 2) MLP: adopting a multi-layer perceptron to aggregate $V^{'}$ and $C^{'}$, 3) MHA: adopting a multi-head attention (MHA) to aggregate $V^{'}$ and $C^{'}$, and 4) GATE: adopting our proposed gate controller to aggregate $V^{'}$ and $C^{'}$. The results are summarized in Tab.~\ref{tab:abl2}. Compared with other control strategies, our method can achieve better performance on all metrics by a large margin. It indicates the effectiveness of our proposed content gate in CIG.

\subsection{Generalization on VideoQA}

To further prove the generalization of our method, we apply proposed video dictionary to video question answering (VideoQA). Tab.~\ref{tab:ab5} shows the experimental results. Practically, we firstly choose three popular VideoQA methods, including HME \cite{fan2019heterogeneous}, MASN \cite{seo2021attend}, and HCRN \cite{HCRN}, as our baseline models. We reproduced their results by running the available code. Then, we simply integrate the video dictionary into the three models via the same operation of VCS. As we can see, our approach all gains a certain level of improvement on the current VideoQA models. For instance, when equipped with our video dictionary, the current SOTA method HCRN can boost the accuracy by around 1.2\% on MSVD-QA and 0.7\% on MSRVTT-QA respectively. Hence, it demonstrates that our proposed visual commonsense has a strong generalization ability in other video-related tasks.

\subsection{Qualitative Results}
Fig.~\ref{fig:vis} illustrates the generated captions on MSR-VTT datasets. Overall, it is observed that the context of captions generated by our model VCRN is more diverse and richer than the baseline model, and involves more activity associations and detailed information. For instance, the example at the top-left shows the baseline model can only understand the general meaning of the video, \ie the descriptions are ``talking about something''. By contract, our model VCRN can recognize more detailed objects and activities (``shirt'', ``candle'', ``soda'' and ``science experiment"). The rest of the examples have similar properties.


Moreover, to better understand the effectiveness of our proposed visual commonsense knowledge, we visualize the attention map between videos and a partial video dictionary in a batch, as illustrated in Fig.~\ref{fig:heatmap}. For example, the first video mainly attends to sports and crowd-related concepts, while the second video contains the ocean, which tends to focus on water-related content highly. It demonstrates that our model can effectively associate the current video with visual commonsense.

\section{Conclusion}
\label{conclusion}
In this paper, we present a novel Visual Commonsense-aware Representation Network (VCRN) for video captioning, which is to mine the cognitive power of the model's visual commonsense knowledge. By constructing a video dictionary from all videos in the dataset, we can obtain effective visual commonsense representation for captioning. Furthermore, our proposed visual concept selection and conceptual integration generation are able to capture video-related commonsense information and generate more accurate captions, respectively. Our proposed model achieves state-of-the-art performance on both MSVD and VATEX datasets and comparable results on MSR-VTT dataset. Extensive experiments and qualitative results have demonstrated the effectiveness of each module. Besides, we also demonstrate the strong generalization of our method by transferring to the video question answering task.




 
\bibliography{cite}

\bibliographystyle{IEEEtran}



 




\vfill

\end{document}